# Unsupervised Domain Adaptation using Graph Transduction Games

Sebastiano Vascon
*DAIS & ECLT*
Ca' Foscari University of Venice
Venice, Italy
sebastiano.vascon@unive.it

Sinem Aslan
*ECLT*
Ca' Foscari University of Venice
Venice, Italy
siinem@gmail.com

Alessandro Torcinovich
*DAIS & ECLT*
Ca' Foscari University of Venice
Venice, Italy
ale.torcinovich@unive.it

Twan van Laarhoven
*Department of Management, Science and Technology*
Open University of the Netherlands
Heerlen, The Netherlands
mail@twanvl.nl

Elena Marchiori
*ICIS*
Radboud University Nijmegen
Nijmegen, The Netherlands
elenam@cs.ru.nl

Marcello Pelillo
*DAIS & ECLT*
Ca' Foscari University of Venice
Venice, Italy
pelillo@unive.it

*Abstract*—Unsupervised domain adaptation (UDA) amounts to assigning class labels to the unlabeled instances of a dataset from a target domain, using labeled instances of a dataset from a related source domain. In this paper we propose to cast this problem in a game-theoretic setting as a non-cooperative game and introduce a fully automated iterative algorithm for UDA based on graph transduction games (GTG). The main advantages of this approach are its principled foundation, guaranteed termination of the iterative algorithms to a Nash equilibrium (which corresponds to a consistent labeling condition) and soft labels quantifying uncertainty of the label assignment process. We also investigate the beneficial effect of using pseudo-labels from linear classifiers to initialize the iterative process. The performance of the resulting methods is assessed on publicly available object recognition benchmark datasets involving both shallow and deep features. Results of experiments demonstrate the suitability of the proposed game-theoretic approach for solving UDA tasks.

## I. INTRODUCTION

The success of deep learning in computer vision classification tasks relies on the availability of a large amount of images annotated with their ground truths. However, manual label annotation is typically an expensive process and may contain wrong annotations. In order to overcome these limitations, Semi-Supervised Learning (SSL) approaches have been developed, usually involving the training of a classifier from a large dataset with plenty of unlabeled data and substantially less annotated data. In some cases however, it is expensive to obtain unlabeled data too, resorting instead on data coming from a different source. This problem is best formulated in the Unsupervised Domain Adaptation setting, where unlabeled data comes from a different related distribution than that of the labeled data. Specifically, an annotated source dataset is exploited to infer the labels of an unlabeled target dataset from a different, related domain. Due to the tight relation between SSL and UDA problems, it is not uncommon to approach them with similar techniques (cf. [1] and [2] for example).

In this paper we investigate the use of a game-theoretic graph-transductive approach, known as Graph Transduction Games, for domain adaptation (GTDA), which has been successfully applied in SSL tasks such as in [3], [4], [5] and [6], and we show that this approach, paired with a preprocessing step, provides overall improvements in three standard domain adaptation cases. Specifically we propose a fully automated pipeline to perform UDA with GTG, comparing our results with those of recent methods, trying also to include prior information provided by a simple classifier, i.e. Logistic Regression. We perform also a comparison of GTG with other three standard graph-transductive algorithms. The picture that arises from the experimental results is promising and suggests considering graph transduction as a key-module when addressing UDA problems. The choice of using GTG as a transduction algorithm for DA has been motivated by its theoretical properties which guarantee: *i*) a consistent labeling of unknown samples at convergence, *ii*) the output of soft-labelings (probability distribution over the classes) for further refinements, *iii*) the possibility of injecting prior knowledge at the beginning of the transductive process.

The main contributions of this work are the following:

- We adopt the theory of label consistency of graph transduction games to propose a principled technique for UDA. This will offer a novel perspective on the UDA problem.
- We propose a parameter-free method for UDA based on game theory which bypasses intensive training phase.
- We reach state-of-the-art performance results on publicly available object recognition domain adaptation tasks.

The paper is organized as follows. To make the paper self-contained, in Sec. II we introduce the GTG algorithm. In Sec. III we describe in detail the proposed method. In Sec. IV we discuss the experimental setting and the competing methods, while in Sec. V we report and analyze our experimental results. Finally, Sec.VI concludes the paper.

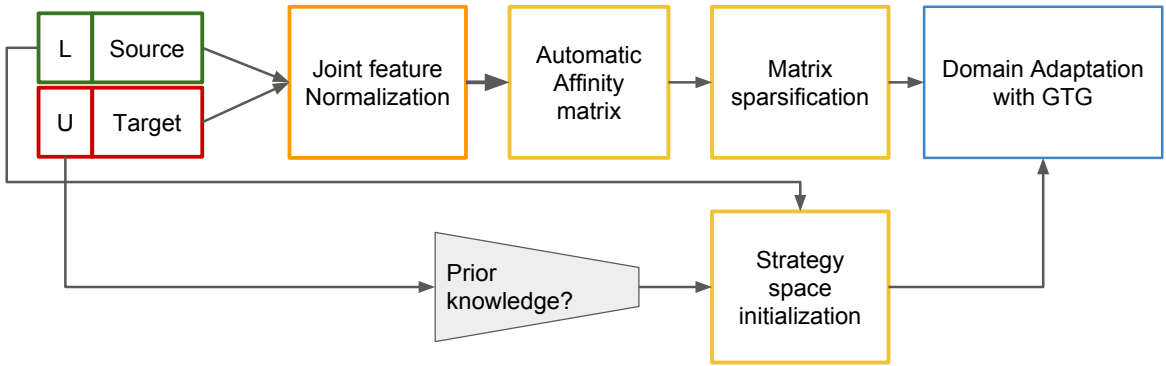

Fig. 1: Pipeline of the proposed method.

*A. Related work*

Many methods for UDA have been introduced. Here we focus on the recent approaches used in our comparative assessment.

A number of DA models align the distributions of features from source and target domains by reducing their discrepancy. For instance, CORrelation ALignment (CORAL) [7] finds a linear transformation that minimizes the distance between the covariance of source and target. Subspace Alignment (SA) [8] computes a linear map that minimizes the Frobenius norm of the difference between the source and target domains, which are represented by subspaces described by eigenvectors. Feature Level Domain Adaptation (FLDA) [9] models the dependence between the two domains by means of a feature-level transfer model that is trained to describe the transfer from source to target domain. FLDA assigns a data-dependent weight to each feature representing how informative it is in the target domain. To to do it uses information on differences in feature presence between the source and the target domain.

Recently, end-to-end UDA methods based on deep neural networks have been shown to perform better than the aforementioned approaches. However they need large train data [10], use target labels to tune parameters [11] and are sensitive to (hyper-)parameters of the learning procedure [12]. Therefore, current state of the art based on this approach start from pre-trained network architectures. Various state-of-the-art methods considered in our comparative analysis use the ResNet pre-trained network, like Deep Domain Confusion (DDC) [13], Deep Adaptation Network (DAN) [14], Residual Transfer Networks (RTN) [11], Reverse Gradient (RevGrad) [12], [15], and Joint Adaptation Networks (JAN) [16].

## II. GRAPH TRANSDUCTION GAMES

Unlike supervised learning methods, semi-supervised learning ones perform classification by leveraging the information coming from both the labeled and unlabeled observations. Their strength appears evident in those cases where the labeled set is not statistically significant to train an accurate fully-supervised classifier. In particular, the sub-category of Graph-Transductive (GT) algorithms perform the classification task in a "transductive reasoning" fashion: rather than trying to infer an inductive, general classification rule, GT methods focus just on the observations at hand (cf. [17] for a more detailed analysis). This lack of generalization can be easily overcome through the training of a supervised classifier with the newly labeled dataset (cf. [3]). In GT the data is modeled as a graph $G = (V, E, w)$ whose vertices are the observations in a dataset and edges represent similarities among them. $V = L \cup U$, where $L = \{(f_1, y_1), ..., (f_l, y_l)\}$ is the set of labeled observations, while $U = \{f_{l+1}, ..., f_n\}$ is the set of unlabeled ones, with $f_i$ being feature vectors and $y_i \in \{1, 2..., m\}$ being labels. Finally, $w$ a general measure of consistency among a pair of observations assigned to each edge $(u, v) \in E$, which can be given in advance or computed using the features of the observations.

*A. Game theoretic framework*

Among the GT algorithms present in the literature, we decided to follow the approach proposed in [18], where the authors interpret the graph transduction task as a non-cooperative multiplayer game.

In Graph Transduction Games (GTG), the graph assumes a second role and represents a context in which a non-cooperative multiplayer game is performed. The players (observations) play games with their neighbors choosing among a set of strategies (labels) and receive a payoff based on the similarities among the neighbors along with the strategies they have chosen. This game is run multiple times until a point of convergence is reached, namely the *Nash Equilibrium* [19].

More formally, we define a set of players $\mathcal{I} = \{1, ..., n\}$ and a set of pure strategies $S = \{1, ..., m\}$ shared among all players (more general settings can be developed as well where players might have different label sets). Each player is associated with a *mixed strategy* $x_i$, which is a probability distribution over $S$ and is intended to model the uncertainty in choosing one pure strategy over another. The mixed strategies reside in the $m$-dimensional *standard simplex* $\Delta^m$ defined as:

$$\Delta^m = \left\{ x_i \in \mathbb{R}^m \ \middle|\ \sum_{h=1}^{m} x_{ih} = 1, x_{ih} \geq 0 \right\} \quad (1)$$

and the mixed strategies of all players compose a *mixed strategy profile* $x \in \Delta^{n \times m}$, residing in the so-called *mixed strategy space*. Pure strategies map directly to *extreme mixed strategies* $e_i^{(h)}$, with 1 at position $h$ and 0 elsewhere. To evaluate the best choice for each player, a tuple of *payoff functions* $u = (u_1, \ldots, u_n)$ s.t. $u : \Delta^{n \times m} \to \mathbb{R}_{\geq 0}^n$ is defined. These payoffs quantify the gain that each player obtains given the actual configuration of the mixed strategy space. It is worth stressing the fact that the payoff functions take into account the mixed strategy of all players, thus fitting well in the GT context.

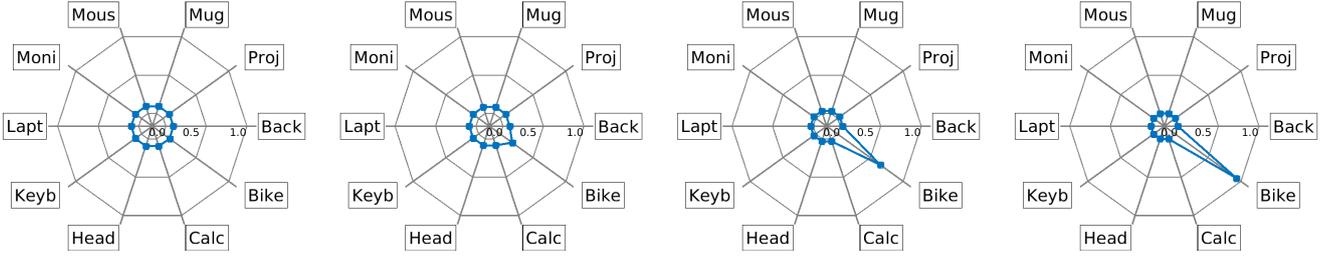

Fig. 2: From left to right the starting point of the dynamical system and the point of convergence. Evolution of the mixed strategy associated to a player during the GTG process. As the dynamic is iterated, the entropy progressively decreases and the distribution peaks toward the correct class.

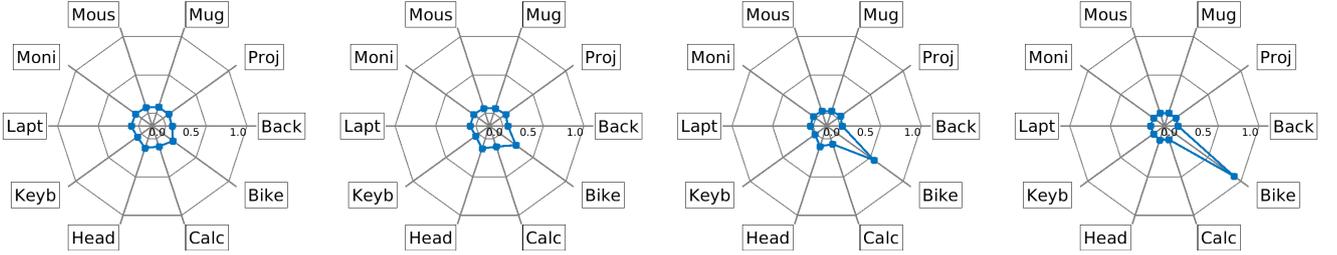

Fig. 3: From left to right the starting point of the dynamical system and the point of convergence. In this example the dynamics start from three different classes, while in the end, thanks to the refinement of the neighboring mixed strategies the correct class is chosen.

Among all the possible configurations we assume to play a polymatrix game [20] where the payoffs associated to each player are additively separable. Without losing generality, let $(e_i^{(h)}, x_{-i})$ define a mixed strategy space where all players $j \in \mathcal{I} \setminus \{i\}$ play their mixed strategy $x_j$ while player $i$ plays the extreme mixed strategy $e_i^{(h)}$, instead. Then:

$$u_i(e_i^{(h)}, x_{-i}) = \sum_{j \in U} (A_{ij} x_j)_h + \sum_{k=1}^{m} \sum_{j \in L} A_{ij}(h, k) \quad (2)$$

$$u_i(x) = \sum_{j \in U} x_i^T A_{ij} x_j + \sum_{k=1}^{m} \sum_{j \in L} x_i^T (A_{ij})_k \quad (3)$$

where $A_{ij} \in \mathbb{R}^{m \times m}$ is the *partial payoff matrix* between the pair of players $(i, j)$. In particular, $A_{ij} = I_m \times \omega_{ij}$ with $I_m$ being the identity matrix and $\omega_{ij}$ the similarity of player $i$ and $j$.

### B. Running GTG

GTG aims at finding an equilibrium condition in the aforementioned game setting, pushing the mixed strategies of each player toward extreme ones. This, in turn, corresponds to having a consistent labeling of the observations [21]. The search for the so-called *Nash equilibria* [19] is achieved through a dynamical system belonging to the class of *Replicator Dynamics* [22] first theorized in [23]. Under this context, the mixed strategies together compose a multi-population where only the fittest pure strategies are kept while the other go extinct.

In our case, we adopted the discrete version of the dynamics:

$$x_{ih}^{(t+1)} = x_{ih}^{(t)} \frac{u_i(e_i^{(h)}, x_{-i}^{(t)})}{u_i(x^{(t)})} \quad (4)$$

where $t$ defines the current iteration of the process. As for the labeled observations, $x_i^{(0)}$ can be simply set to the extreme mixed strategy corresponding to $y_i$, namely $x_{ih} = \mathbb{1}(h = y_i)$, where $\mathbb{1}$ is the indicator function, while for the unlabeled ones they can be set either with some prior or a uniform distribution, thus $x_{ih}^{(0)} = 1/m$, $\forall h \in S$ (we tried both cases in our experiments). The dynamics are run until two consecutive steps do not differ significantly or a maximum number of iteration is reached. In Fig. 2 and Fig. 3 we show two examples of the GTG algorithm in the case of absence of prior knowledge and in its presence, respectively.

## III. Domain Adaptation with GTG

In this section we present our method, GTDA. We will explain how to cast the unsupervised domain adaptation problem in graph-transduction game setting. We consider given labeled ($L$) and unlabeled ($U$) datasets from the source and target domain, respectively. Then, labels from source data are propagated to the target instances by playing a non-cooperative multiplayer game in which the players are the objects (instances) and the labels the possible strategies.

The interaction between the players are represented in terms of a weighted undirected graph in which the edges are weighted by the similarity of player pairs, hence how much they will affect each others. In particular, the process is illustrated in Fig. 1 and explained in the following steps:

*a) Joint feature standardization:* Given a dataset of features $D$ and two domains $d_s, d_t \in D$ (source and target respectively), we normalize their features jointly as a pre-processing step. We perform two types of normalization on the union of the features: *std* features are scaled by their standard deviation or *z-score* subtract the mean and scaled by their standard deviation. Depending whether the sparsity of the features should be preserved or not, we pick the *std* or *z-score*, respectively.

*b) Initialization of the mixed strategy profile:* The initial mixed strategy profile of the players, denoted as $x^{(0)}$ represents the starting point of the game. If prior knowledge is available, we can leverage it for its initialization. In our experimental settings, we explore two different initializations: *i)* in which no prior information is exploited (*no-prior* in the following) and *ii)* where an output from a logistic regression classifier is used (*+LR*). In the latter case, the logistic regression classifier has been trained for each pair of jointly normalized domains in a dataset. The training has been performed considering only the features belonging to the source, in a 2-fold cross validation setting, with an hyperparameter search for the $C$ variable in the following log-scale range $C = [10^{-3}, 10^4]$. We end up with a LR model $M_{i,j}$ for each pair of domains $d_i$ and $d_j$ in a dataset. Given an unlabeled observation, the LR model outputs a probability distribution over the classes which is later used as prior knowledge in the strategy space. The choice of having as prior a probability distribution for each unlabeled object, instead of a one-hot vector, is mandatory since the one-hot vector cannot be updated by the GTG algorithm hence the performances would have been the same as the LR itself.

**Algorithm 1** GTDA algorithm
***
**Input:** source feature matrix $F_S \in \mathbb{R}^{|S| \times d}$, target feature matrix $F_T \in \mathbb{R}^{|T| \times d}$, one-hot source labels $Y_S \in \Delta^{|S| \times m}$, minimum tolerance $\varepsilon$, maximum number of iterations $K$.
**Output:** target soft predictions $Y_T \in \Delta^{|T| \times m}$
1: $N = |S| + |T|$
2: $\hat{F}_S = \text{normalize}(F_S)$ ▷ Sec. III.a
3: $\hat{F}_T = \text{normalize}(F_T)$ ▷ Sec. III.a
4: $P_T = \text{LR}(\hat{F}_S, Y_S, \hat{F}_T)$ ▷ Get log. reg. priors for $F_T$
5: $x(0) = \begin{bmatrix} Y_S \\ \hat{P}_T \end{bmatrix}$ ▷ Init. Mixed Strategy Prof. Sec. III.b
6: $W = [\omega(i,j)]_{ij}$ ▷ Eq. 5
7: $\hat{W} = \text{sparsify}(W)$ ▷ Sec. III.d
8: $tol = +\infty$, $t = 0$
9: **while** $tol \geq \varepsilon$ and $t < K$ **do**
10:     **for** $i = 1, \ldots, N$ **do**
11:         $x_i(t+1) = \frac{x_i(t) \odot (\hat{W}x(t))_i}{x_i(t)(\hat{W}x(t))_i^T}$ ▷ Eq. 4
12:     $tol = \|x(t+1) - x(t)\|_2$
13:     $t = t + 1$
14: $Y_T = x(t-1)_{|S|:N, 1:m}$ ▷ Get the target predictions

*c) Computation of the affinity matrix:* The core of the GTG is stored in the affinity matrix $W$ (payoff of the players), so its computation requires particular care. In our experimental setting, we decided to rely on the following standard similarity kernel:

$$\omega(i,j) = \begin{cases} \exp\left\{-\frac{d(f_i, f_j)^2}{\sigma_i \sigma_j}\right\} & \text{if } i \neq j \\ 0 & \text{else} \end{cases} \quad (5)$$

where $f_i$, $f_j$ are the features of observations $i$ and $j$ respectively, $d(f_i, f_j)$ is the cosine distance between features $f_i$ and $f_j$. Here, motivated by [24] and [5], we set the scaling parameter $\sigma_i$ automatically, considering the local statistics of the neighborhood of each point. Accordingly to [24], the value of $\sigma_i$ is set to the distance of the 7-th nearest neighbour of observation $i$.

*d) Affinity sparsification:* The sparsification of the graph plays an important role in the performances of the algorithm. Indeed, filtering out the small noisy similarities which may bias the utilities in Eq. 5, would prevent incorrect class labelings.

Here, we follow a statistical connectivity principle in random graph, which states that a graph is connected if each node has at least $k = \lfloor log_2(n) \rfloor + 1$ nearest neighbours [25]. The rationale of this choice is that the labels in GTG are propagated from the labeled elements to the unlabeled ones. If the graph is not connected the propagation might get stuck at a certain point. This sparsification ensures that the graph is connected, hence all the nodes are reached at the equilibrium condition of the dynamical system (cf. Eq4). The sparsification is performed for each node $i$ independently considering the distance value of the $k$-NN as a threshold for the other nodes in the graph. In order to obtain a symmetric neighborhood, we include the node $i$ in the neighbourhood of $j$ (and viceversa) if one of the two is in the neighborhood of the other.

*e) Execution of GTG:* Once the affinity matrix is computed and the mixed strategy profile is initialized, GTG can be finally played up to convergence. The final probabilities, which determine then the labels for the unlabeled observations, correspond to the adaptation from sources to targets.

In algorithm 1 we present the pseudo-code of the entire method.

## IV. EXPERIMENTAL SETTING

To validate our approach, we perform experiments on two publicly available popular datasets for object recognition domain adaptation: the Office-Caltech 10 [26] and the Office 31 [27].

### A. Datasets

In the following we present a short description of each datasets used for our experiments.

*a) Office-31:* Office-31 [27] is a dataset containing 31 classes divided in 3 domains: *Amazon* (A), *DSLR* (D) and *Webcam* (W). Office-31 has a total of 4110 images, with a maximum of 2478 images per domain. In this dataset we use deep features extracted from the ResNet-50 architecture [28] pretrained on ImageNet.

*b) Office-Caltech:* Office-Caltech [26] consists in observations taken from the common classes of Office 31 and Caltech256 (10 in total) and are divided in 4 image domains, namely the ones of Office-Caltech and the additional *Caltech* (C). The features we consider are of two kinds: 800 SURF features [29], which we preprocess by z-score standardization, and deep features in the same fashion as the previous dataset.

### B. Evaluation Criteria

We evaluate and report the accuracy on the target domain for each adaptation. Accuracy is computed as the fraction of the correctly labeled target instances. Furthermore, we report the average accuracy per methods and the top-3 performing by different coloring (**best performing**, second and third). Along the analysis of the results we highlight also the number of hit time that a method perform better than the competitors.

### C. Comparing Methods

In order to assess our method in a broad context, the performances of GTDA has been compared with both standard domain adaptation methods, recent deep-learning based algorithm and baselines classifiers (SVM and LR). Furthermore, we assess the effectiveness of GTG, replacing it with other GT methods (Label Propagation, Label Spreading and Gaussian Fields and Harmonic Functions).

In our experiments, since we are dealing with more than two classes, we used one-vs-all linear SVM and multi-class logistic regression. More details on the methods we experimented for comparative analysis are given below.

*1) Shallow Domain Adaptation Methods:* The most prevalent domain adaptation methods accomplish domain adaptation task by reducing the discrepancy between source and target distributions via computing a feature transformation. We chose CORrelation ALignment (CORAL) [7] and Subspace Alignment (SA) [8] which are two popular methods following this approach. Reported performances for both methods are appealing whereas their application to high dimensional data might be problematic since they are not scalable to high number of features. Another approach for domain adaptation is modeling the dependence of source and target domain in feature level. We experimented by a recent work, namely Feature Level Domain Adaptation (FLDA) [9], that follows this approach. We use published source codes of the shallow DA methods for all datasets.

*2) DNN-based Domain Adaptation Methods:* Motivated by their reported stunning performances in recent years, we compared performance of GTDA with the performances of a number of Deep Neural Networks-based domain adaptation methods reported on the Office 31 dataset based on ResNet50 features. Specifically, we make comparison with Deep Domain Confusion (DDC) [13] where an objective function including an additional domain confusion term is proposed for learning domain-invariant representations for classification, Deep Adaptation Network (DAN) [14] where more transferable features are learned by adapting source and target distributions in multiple task-specific layers, Residual Transfer Networks (RTN) [11] that achieves feature adaptation and classifier adaptation simultaneously by deep residual learning [28], Reverse Gradient (RevGrad) [12], [15] that improves domain adaptation by employing adversarial training paradigm, and Joint Adaptation Networks (JAN) [16] that uses an adversarial learning strategy to maximize a joint maximum mean discrepancy such that distributions of source and target domains be more distinguishable. Despite of high performance accuracies, some disadvantages of DNN-based methods to be taken into account are that they require abundant training data for improvement in performance, they use target labels for parameter tuning [11] and their sensitivity to hyperparameters of the learning procedure is high [12]. We refer to the performance results reported in [16] for the aforementioned DNN-based techniques to make comparison with our technique, hence we will add results for the Office31 dataset only.

*3) Graph Transduction Techniques for Domain Adaptation:* We compare our game-theoretic graph transduction technique against three other transductive techniques, namely Label Propagation (LP) [30], Label Spreading (LS) [31] and Harmonic Function (HF) [32] for the domain adaptation problem. Similar to our method, these techniques exploit the so-called smoothness principle which states that closer instances tend to belong to the same class. LP [30] performs hard clamping of input labels which yields to avoiding change on the original label distribution at every iteration, while LS [31] adopts soft clamping where initial assignments are changed by a fraction $\alpha$ at each iteration. Moreover, employing regularization, the cost employed in LS differentiates from LP, which provides better robustness to noise. *Gaussian Fields and Harmonic Functions (HF)* [32] tries instead to compute a function $f$ by minimizing a corresponding energy function $E(f)$. The solution is *harmonic* and this property can be exploited to propagate information according to the aforementioned smoothness principle.

TABLE I: Comparative analysis on Office-31 dataset (ResNet-50 features)

|  | A→D | A→W | D→A | D→W | W→A | W→D | avg |
|---|---|---|---|---|---|---|---|
| **Baselines** | | | | | | | |
| Source SVM | 76.9 | 73.8 | 60.3 | 97.5 | 59.4 | **100.0** | 78.0 |
| Source LR | 74.7 | 70.8 | 60.6 | 97.5 | 60.2 | **100.0** | 77.3 |
| **Shallow models** | | | | | | | |
| SA | 76.7 | 75.5 | 62.2 | 97.9 | 60.3 | **100.0** | 78.8 |
| FLDA-Q | 76.3 | 75.5 | 59.9 | 97.5 | 58.6 | 99.8 | 77.9 |
| CORAL | 78.9 | 76.9 | 59.7 | 98.2 | 59.9 | **100.0** | 78.9 |
| **Graph-transductive methods** | | | | | | | |
| Lab Prop | 2.4 | 3.6 | 3.3 | 3.6 | 3.3 | 99.8 | 19.3 |
| Lab Spread | 77.3 | 79.2 | 63.1 | 98.6 | 60.8 | 99.8 | 79.8 |
| Harmonic Function | 73.7 | 80.3 | 62.3 | 98.1 | 46.8 | 99.8 | 76.8 |
| **Proposed method, GTDA** | | | | | | | |
| GTDA | 80.5 | 78.0 | 66.2 | **98.9** | 62.9 | 99.8 | 81.1 |
| GTDA + LR | **82.5** | **84.2** | **67.1** | 97.9 | **69.1** | 99.8 | **83.4** |

TABLE II: Comparative analysis on Office-31 dataset (ResNet-50 features)

|  | A→D | A→W | D→A | D→W | W→A | W→D | avg |
|---|---|---|---|---|---|---|---|
| **Deep Neural Networks (results taken from [16])** | | | | | | | |
| DDC | 76.5 | 75.6 | 62.2 | 96.0 | 61.5 | 98.2 | 78.3 |
| DAN | 78.6 | 80.5 | 63.6 | 97.1 | 62.8 | 99.6 | 80.4 |
| RTN | 77.5 | 84.5 | 63.6 | 96.8 | 64.8 | 99.4 | 81.6 |
| RevGrad | 79.7 | 82.0 | 68.2 | 96.9 | 67.4 | 99.1 | 82.2 |
| JAN-A | **85.1** | **86.0** | **69.2** | 96.7 | **70.7** | 99.7 | **84.6** |
| **Proposed method, GTDA** | | | | | | | |
| GTDA | 80.5 | 78.0 | 66.2 | **98.9** | 62.9 | **99.8** | 81.1 |
| GTDA + LR | 82.5 | 84.2 | 67.1 | 97.9 | 69.1 | **99.8** | 83.4 |

To make a fair comparison with our approach, we provide to these algorithms the same affinity matrix as our, i.e. $W$, which is computed using the same scheme of normalization, $\sigma$ selection and sparsification. For the LS technique, we experimented with a variety of values in the range of $(0, 1)$ for the parameter $\alpha$. Since we got the best results when $\alpha = 0.2$ which is also the suggested default value we present the results of LS with $\alpha = 0.2$.

## V. RESULTS

We use the notation of $A \to B$ to indicate the adaptation with $A$ as source and $B$ as target dataset. While we discuss the performances of the techniques, (i) we consider the averaged accuracy over all adaptation tasks and (ii) the number of adaptation tasks that a method outperforms.

### A. Office 31 dataset

We present performances on Office 31 for the shallow and graph transductive methods in Table I while additional comparisons with deep-learning models is outlined in Table II reporting results from [16].

*a) Non DL methods:* The results on non-DL methods are reported in Table I. The GTDA outperforms CORAL, i.e. the best performed shallow DA method on this dataset, by around 2% and 4% in averaged accuracy with and without prior, respectively. When we compare GTDA with other GT methods, i.e. Label Prop., Label Spread and Harmonic Function, we see that GTDA without prior outperforms all of them with a performance of 81.1%, while Label Spread follows GTDA with the performance of 79.8 %. When the prior knowledge is used in GTDA the performances are far better, being the top performing ones. Another point to highlight from this experiment is that, in general, the transductive methods outperform the shallow models. Without considering the average results, the GTDA+LR outperforms 5 over 6 times the shallow models. When prior is not used the GTDA is the second best performing algorithm (still considering the GTDA+LR).

*b) DL-based methods:* The results on DL-based methods are reported in Table II. The GTDA outperforms DDC, DAN, RTN and RevGrad which are end-to-end learned system for DA. This is surprising, considering that GTDA does not require an extensive training phase and neither a parameter optimization like in DNN. JAN-A achieves the best averaged accuracy, i.e. 84.6%, on this dataset being for 4 times the best performing and just once as a second. Among the other DL models no one is able to clearly outperform JAN-A. Our GTDA without prior outperform JAN-A on 2 cases (D-W, W-D) while in the remaining five it becomes as third best only once. When GTDA benefits from prior (GTDA +LR) the performances approaches to JAN-A (83.4 % vs. 84.6 %) becoming the second best algorithm, even outperforming the other DL approaches.

### B. Office-Caltech 10 dataset

We present performances on this dataset in Tables III and IV when SURF and ResNet50 features are used, respectively. In Table III we see that CORAL outperforms other shallow DA methods, i.e. FLDA and Source SVM, by achieving the second best averaged accuracy over all the methods. While we achieve almost same performance as CORAL on average (48.8% for CORAL and 48.5% for GTDA) when we do not use priors. When we get benefit of the priors (GTDA+LR) we outperform CORAL by 1% in average accuracy becoming the best one over all methods. We outperform other GT methods both in averaged accuracy and at majority of the adaptation tasks. In particular, our best competitor is CORAL reaches 5 top results among the shallow models while GTDA with prior outperform 6 time the shallow ones becoming the more stable algorithm in this setting.

We see in Table IV that when ResNet50 features are used the performances of all methods are improved significantly over the ones obtained when SURF features were used, except Label Prop. All shallow DA methods and GTDA (when not using prior) achieve very similar performances. In particular, while FLDA-Q and CORAL outperforms GTDA in averaged accuracy when prior was not used by 0.2% and 0.1%, we see that GTDA reaches to 3 top results while FLDA-Q and CORAL stay at 1 and 2, respectively. Following them, baseline methods and Lab Spread achieves similar performances in averaged accuracy. When we get benefit from prior we outperform shallow DA methods at every adaptation task (except $C \rightarrow W$), i.e. we reach to the best result at 8 adaptation tasks, second best results at 3 adaptation tasks and third best result at 1 adaptation task, and we are the best among both shallow DA and other GT methods with 94.8% in averaged accuracy with GTDA +LR.

TABLE III: Comparative analysis on Office-Caltech dataset (SURF features)

| | A→C | A→D | A→W | C→A | C→D | C→W | D→A | D→C | D→W | W→A | W→C | W→D | avg |
|---|---|---|---|---|---|---|---|---|---|---|---|---|---|
| Baselines | | | | | | | | | | | | | |
| Source SVM | 41.0 | 40.1 | 42.0 | 52.7 | 45.9 | **47.5** | 33.0 | 32.1 | 75.9 | 38.4 | 34.6 | 75.2 | 46.5 |
| Source LR | 42.8 | 36.3 | 35.3 | **54.1** | 42.7 | 40.7 | 33.9 | 31.2 | 83.1 | 37.3 | 32.9 | 71.3 | 45.1 |
| Shallow Models | | | | | | | | | | | | | |
| SA | 37.4 | 36.3 | 39.0 | 44.9 | 39.5 | 41.0 | 32.9 | 34.3 | 65.1 | 34.4 | 31.0 | 62.4 | 41.5 |
| FLDA-L | 41.5 | **45.9** | 42.0 | 49.5 | **48.4** | 44.1 | 31.7 | 34.1 | 75.6 | 35.3 | 33.8 | 72.6 | 46.2 |
| FLDA-Q | 43.5 | 43.3 | 40.7 | 53.5 | 44.6 | 45.1 | 30.8 | 31.2 | 73.2 | 35.2 | 32.1 | 75.8 | 45.7 |
| CORAL | **45.1** | 39.5 | **44.4** | 52.1 | 45.9 | 46.4 | 37.7 | 33.8 | 84.7 | 35.9 | 33.7 | 86.6 | 48.8 |
| Graph-transductive methods | | | | | | | | | | | | | |
| Lab Prop | 13.4 | 7.6 | 9.8 | 9.6 | 45.9 | 9.8 | 9.6 | 13.4 | 9.8 | 9.6 | 13.4 | **89.8** | 20.2 |
| Lab Spread | 41.3 | 36.3 | 32.5 | 53.3 | 47.8 | 41.4 | 36.1 | 34.2 | 90.2 | 36.0 | 34.2 | 88.5 | 47.7 |
| Harmonic Function | 41.1 | 38.9 | 35.9 | 52.2 | 47.1 | 37.6 | 30.8 | 29.3 | 89.2 | 32.2 | 32.7 | 88.5 | 46.3 |
| Proposed method, GTDA | | | | | | | | | | | | | |
| GTDA | 40.2 | 37.6 | 32.9 | 53.8 | 46.5 | 35.9 | **41.3** | **39.9** | 92.2 | 34.6 | **38.5** | 89.2 | 48.5 |
| GTDA + LR | 40.2 | 37.6 | 38.3 | 52.6 | 45.9 | 45.1 | 39.2 | 35.4 | 92.2 | **41.0** | 37.1 | 89.2 | **49.5** |

TABLE IV: Comparative analysis on Office-Caltech dataset (ResNet-50 features)

| | A→C | A→D | A→W | C→A | C→D | C→W | D→A | D→C | D→W | W→A | W→C | W→D | avg |
|---|---|---|---|---|---|---|---|---|---|---|---|---|---|
| Baselines | | | | | | | | | | | | | |
| Source SVM | 91.0 | 88.5 | 87.5 | 94.1 | 94.9 | 87.8 | 90.0 | 86.1 | 98.6 | 89.1 | 85.9 | 100.0 | 91.1 |
| Source LR | 89.9 | 91.7 | 88.5 | 94.5 | 93.6 | 85.1 | 90.1 | 85.8 | 98.0 | 89.7 | 85.5 | 100.0 | 91.0 |
| Shallow models | | | | | | | | | | | | | |
| SA | 89.7 | 93.0 | 90.8 | 94.6 | 91.1 | 93.2 | 89.8 | 84.1 | 99.0 | 88.9 | 84.3 | 100.0 | 91.5 |
| FLDA-Q | 91.1 | 93.6 | 92.2 | 94.5 | 94.3 | 89.5 | 90.3 | 86.3 | 97.6 | 90.3 | 83.7 | 100.0 | 91.9 |
| CORAL | 85.9 | 91.1 | 89.8 | 94.3 | 93.0 | **93.2** | 92.8 | 86.8 | 98.6 | 90.9 | 85.5 | 100.0 | 91.8 |
| Graph-transductive methods | | | | | | | | | | | | | |
| Lab Prop | 13.4 | 7.6 | 9.8 | 9.6 | 7.6 | 9.8 | 9.6 | 13.4 | 9.8 | 9.6 | 13.4 | 100.0 | 17.8 |
| Lab Spread | 87.1 | 88.5 | 95.9 | 93.4 | 91.1 | 84.1 | 88.0 | 88.6 | **99.7** | 90.1 | 85.6 | 100.0 | 91.0 |
| Harmonic Function | 88.6 | 80.9 | 85.4 | 93.5 | 94.9 | 89.2 | 88.1 | 83.2 | **99.7** | 57.4 | 69.2 | 100.0 | 85.8 |
| Proposed method, GTDA | | | | | | | | | | | | | |
| GTDA | 90.0 | 87.9 | **98.0** | 93.5 | 91.7 | 79.7 | 89.4 | **89.4** | 99.3 | 93.2 | 88.8 | 100.0 | 91.7 |
| GTDA + LR | **91.5** | **98.7** | 94.2 | **95.4** | **98.7** | 89.8 | **95.2** | 89.0 | 99.3 | **95.2** | **90.4** | 100.0 | **94.8** |

*C. Overall analysis*

The method yields competitive results in different adaptation setting. This is surprising since the method is quite simple and the choice made by the dynamical system are greedy. The other thing that is worth noting is the use of the prior, which significantly improves the performance. Prior knowledge can be easily injected in the model through the strategy space and it is of legitimate use even under unsupervised DA providing that the training for whichever model is performed only on the source data. The proposed GTDA is the most stable in terms of number of best DA, thereby making it an attractive alternative.

## VI. CONCLUSIONS

In this work we have proposed a new algorithm to tackle unsupervised domain adaptation tasks. The methodology is based on graph-transduction and game-theory, offering a principled perspective to the problem. The GTDA proposed here has two main advantages: *i)* it is completely parameter-free and *ii)* it allows the direct embedding of prior knowledge on the target labels to be predicted. The results achieved on publicly available benchmark datasets demonstrate the validity of the proposed approach, whose performance is competitive with respect to state-of-the-art DA techniques with shallow and deep features as well as to other standard graph-based transductive methods. Furthermore, GTDA reaches comparable performances as that of known deep-learning UDA methods. As a future work we plan to extend the comparison with other recent DA techniques using more real-life datasets. As for the methodology we are interested in investigating other aspects, like semi-supervised domain adaptation.

ACKNOWLEDGMENT

This work has been partially funded by the Netherlands Organization for Scientific Research (NWO) within the EW TOP Compartment 1 project 612.001.352.